\documentclass[runningheads]{llncs}
\usepackage{graphicx}
\usepackage{mathtools}
\usepackage{amssymb}
\usepackage{tikz-cd}
\usepackage{xfrac}

\usepackage{color}
\usepackage{xspace}
\usepackage{algorithm2e}
\usepackage{enumerate}
\usepackage{paralist}
\usepackage{amsmath}

\usepackage{subcaption}
\newcommand{\ie}{\textit{i}.\textit{e}.,\xspace}
\newcommand{\eg}{\textit{e}.\textit{g}.,\xspace}
\SetKwInput{KwInput}{Input}
\SetKwInput{KwOutput}{Output}

\DeclareMathOperator*{\argmax}{arg\,max}

\begin{document}
\title{Knowledge Graph Embedding for Ecotoxicological Effect Prediction}
%
%
\author{Erik B. Myklebust\inst{1,2}\thanks{Corresponding author: Erik B. Myklebust, \email{ebm@niva.no}} \and Ernesto Jimenez-Ruiz\inst{2,3} \and Jiaoyan Chen\inst{4} \and \\Raoul Wolf\inst{1} \and Knut~Erik Tollefsen\inst{1}}
\authorrunning{E. B. Myklebust et. al.}
%
\institute{
Norwegian Institute for Water Research, Oslo, Norway \and
Department of Informatics, University of Oslo, Norway \and
Alan Turing Institute, London, United Kingdom \and
Department of Computer Science, University of Oxford, United Kingdom
}

%
\maketitle              
\setcounter{footnote}{0} 
\begin{abstract}

Exploring the effects a chemical compound has on a species takes a considerable experimental effort.  Appropriate methods for estimating and suggesting new effects can dramatically reduce the work needed to be done by a laboratory. 
In this paper we explore the suitability of using a  knowledge graph embedding approach for ecotoxicological effect prediction. 
A knowledge graph has been constructed from publicly available data sets, including a species taxonomy and chemical classification and similarity. The publicly available effect data is integrated to the knowledge graph using ontology alignment techniques. 
Our experimental results show that the knowledge graph based approach improves the 
selected baselines.

\end{abstract}
\section{Introduction}
Extending the scope of risk assessment models is a long-term goal in ecotoxicological research. However, biological effect data is only available for a few combinations of chemical-species pairs.\footnote{Chemical and compound are used interchangeably.} Thus, one of the main efforts in ecotoxicological research is the design of
tools and methods to extrapolate from known to unknown combinations in order to facilitate risk assessment predictions on a population basis.

The Norwegian Institute for Water Research (NIVA) is a
leading Norwegian institute for fundamental and applied research on marine and freshwaters.\footnote{NIVA Institute: \url{https://www.niva.no/en}}
The Ecotoxicology and Risk Assessment programme at NIVA has through the last years developed a 
risk assessment system called RAdb.\footnote{NIVA Risk Assessment Database: \url{https://www.niva.no/en/projectweb/radb}} This system has been applied to several case studies based on agricultural/industrial runoff into lakes or fjords. However, the underlying relational database structure of RAdb has its limitations when dealing with the integration of diverse data and knowledge sources. 
This limitation is exacerbated when these resources 
do not share a common vocabulary, as it is the case in our ecotoxicology risk assessment setting. 

In this paper we present a preliminary study of the benefits of using Semantic Web tools to integrate different data sources and knowledge graph embedding approaches to improve the ecotoxicological effect prediction.  
Hence, our contribution to the NIVA institute is twofold:
\vspace{-0.2cm}
\begin{enumerate}[\it (i)]
    \item We have created a knowledge graph by gathering and integrating the relevant biological effect data and knowledge. Note that the format of the source data varies from tabular data, to SPARQL endpoints and ontologies. In order to discover equivalent entities we exploit internal resources, external resources (\eg Wikidata \cite{wikidata2014}) and ontology alignment (\eg LogMap~\cite{logma_ecai2012}).   
    \item We have evaluated three knowledge graph embedding models (TransE \cite{NIPS2013_5071}, DistMult \cite{Yang2015EmbeddingEA} and HolE \cite{DBLP:journals/corr/NickelRP15}) together with the (baseline) prediction model currently used at NIVA. Our evaluation shows a considerable improvement with respect to the baseline and the benefits of using the knowledge graph models in terms of recall and $F_{\beta=2}$ score. Note that, in the NIVA use case, \textit{false positives} are preferred over \textit{false negatives} (\ie missing the hazard of a chemical over a species).
\end{enumerate}

The rest of the paper is organised as follows. Section 2 provides some preliminaries to facilitate the understanding of the subsequent sections.
In Section 3 we describe the use case where the knowledge graph and prediction models are applied. The creation of the knowledge graph is described in Section 4. Section 5 introduces the effect prediction models, while Section 6 presents the evaluation of these models. Finally, Section 7 elaborates on the contributions and discusses future directions of research.






\section{Preliminaries}

\textbf{Knowledge graphs}. We follow the RDF-based notion of knowledge graphs \cite{j.websem510} which are composed by 
RDF triples  
$\left\langle s, p, o \right\rangle$,
where $s$ represents a subject (a class or an instance), 
$p$ represents a predicate (a property) 
and $o$ represents an object
(a class, an instance or a data value \eg text, date and number).
RDF entities (\ie classes, properties and instances) are represented by an URI (Uniform Resource Identifier).
A knowledge graph can be split into a TBox (terminology), often composed 
by RDF Schema constructors like class subsumption (\eg \texttt{ncbi:taxon/6668}  \texttt{rdfs:subClassOf} \texttt{ncbi:taxon/6657}) and
property domain and range (\texttt{ecotox:affects} \texttt{rdfs:domain} \texttt{ecotox:Chemical}),\footnote{The OWL 2 ontology language provides more expressive constructors. Note that the graph projection of an OWL 2 ontology can be seen as a knowledge graph (\eg \cite{jbs2018}).} and 
an ABox (assertions), which contain relationships among instances (\eg \texttt{ecotox:chemical/330541} \texttt{ecotox:affects} \texttt{ecotox:effect/202}) and semantic type definitions (\eg \texttt{ecotox:taxon/28868} \texttt{rdf:type} \texttt{ecotox:Taxon}). RDF-based 
knowledge graphs
can be accessed with SPARQL queries, the standard language to query RDF graphs.

\medskip
\noindent
\textbf{Ontology alignment}. Ontology alignment is the process of finding mappings or correspondences between a source and a target ontology or knowledge graph~\cite{om2013}. These mappings are typically represented as equivalences among the entities of the input resources (\eg \texttt{ncbi:taxon/13402} \texttt{owl:sameAs} \texttt{ecotox:taxon/Carya}).

\medskip
\noindent
\textbf{Embedding models}. 
Knowledge graph embedding \cite{KGE_survey_2017} plays a key role in link prediction problems where the goal is to learn a scoring function $S:\mathcal{E}\times\mathcal{R}\times\mathcal{E} \to \mathbb{R}$. $S(s,p,o)$ is proportional to the probability that a triple $\left\langle s, p, o \right\rangle$ is encoded as true. Several models 
have
been proposed, \eg Translating embeddings model (TransE) \cite{NIPS2013_5071}. These models are applied to knowledge graphs to resolve missing facts in largely connected knowledge graphs, such as 
DBpedia
\cite{dbpedia2015}. 
Embedding models have also been successfully applied in biomedical link prediction tasks~(\eg \cite{DBLP:journals/bioinformatics/AlshahraniKMKQH17,DBLP:conf/semweb/AgibetovS18}).

\medskip
\noindent
\textbf{Evaluation metrics}. We use (A)ccuracy, (P)recision, (R)ecall, ($F_{\beta}$) score to evaluate the models.
They are defined as 
\begin{align}
    A &= \frac{tp + tn}{tp+tn+fp+fn} \\
    P &= \frac{tp}{tp +fp} \\
    R &= \frac{tp}{tp +fn} \\
    F_{\beta} &= (1+\beta^2)\frac{PR}{\beta^2 P + R}
\end{align}
where $tp$, $tn$, $fp$, and $fn$ stand for \emph{true positive}, \emph{true negative}, \emph{false positive}, and \emph{false negative}, respectively.  Essentially, accuracy is the proportion of correct classifications. Recall is a measure of how many expected positive predictions were found by our model, and precision is the proportion of predictions that were correctly classified.
$F_{\beta}$ is a combined measure of precision and recall. $\beta=1$ gives equal weight, while $\beta < 1$ favours precision and $\beta > 1$ favours recall. 
Here we use $F_{\beta=1}$ ($F_1$ in short) and $F_{\beta=2}$. 

As the above metrics all depend on 
a
selected threshold,
we also use area under the receiver operating characteristic (ROC) curve (AUC) to measure and compare the overall pattern recognition capability of the prediction models. 
ROC is the curve of true positive rate ($\sfrac{tp}{(tp+fn)}$, i.e., recall) and false positive rate ($\sfrac{fp}{(fp+tn)}$), with the threshold ranging from $0$ to $1$ using a small step.
AUC is the area under this curve, its values range between $0$ and $1$.
Larger AUC indicates higher performance.

\section{NIVA use case: ecotoxicology and risk assessment}

Ecotoxicology is a multidisciplinary field that studies the ecological and toxicological effects of chemical pollutants on populations, communities and ecosystems. Risk assessment is the result of the intrinsic hazards of a substance combined with an estimate of the environmental exposure (\ie Hazard + Exposure~=~Risk).

The Computational Toxicology Program within NIVA's Ecotoxicology and Risk Assessment section aims at designing and developing prediction models to assess the effect of chemical mixtures over a population where
traditional laboratory data cannot be easily acquired.

\begin{figure}[tb]
    \centering
    \includegraphics[width=0.8\textwidth]{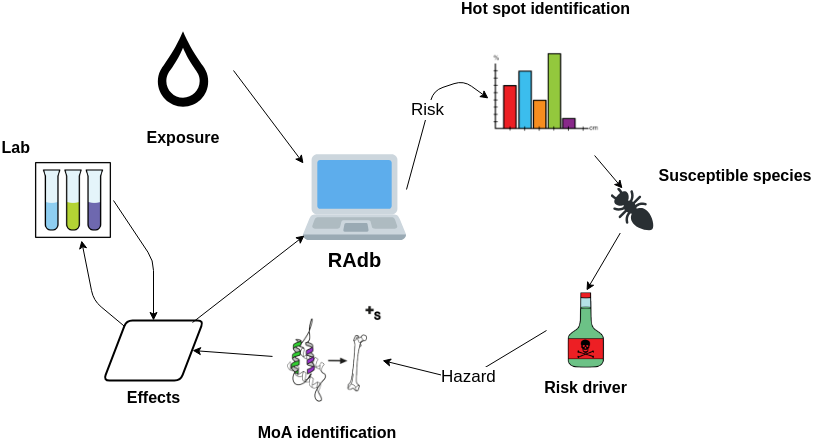}
    \caption{NIVA risk assessment pipeline.}
    \label{fig:niva-pipeline}
\end{figure}

Figure \ref{fig:niva-pipeline} shows the risk assessment pipeline followed at NIVA. %
\textit{Exposure} is data gathered from the environment, while \textit{effects} are hypothesis that are tested in a laboratory. These two data sources are used to calculate risk, which is used to find (further) susceptible species 
and the mode of action (MoA) or type of impact a compound would have over those species. Results from the MoA analysis are used as new effect hypothesis.

\begin{table}[tb]
    \centering
    \begin{tabular}{c|c|l}
        Proportion & Abbreviation & Description \\ \hline
        $0.21$ & NR & Not reported\\
        $0.17$ & NOEL & No-observable-effect-level\\
        $0.16$ & LC50 & Lethal concentration for $50\%$ of test population\\
        $0.14$ & LOEL & Lowest-observable-effect-level\\
        $0.05$ & NOEC & No-observable-effect-concentration\\
        $0.05$ & EC50 & Effective concentration for $50\%$ of test population\\
        $0.04$ & LOEC & Lowest observable effect concentration\\
        $0.03$ & BCF & Bioconcentration factor\\
        $0.02$ & NR-LETH & Lethal to $100\%$ of test population\\
        $0.02$ & LD50 & Lethal dose for $50\%$ of test population \\
        $0.11$ & Other &  
    \end{tabular}
    \medskip
    \caption{The 10 most frequent outcomes in ECOTOX effect data.}
    \label{tab:endpoints}
    
\end{table}

The effect data is gathered during experiments in a laboratory, where the population of a single species is exposed to a concentration of a toxic compound. Most commonly, the mortality rate of the population is measured at each time interval until 
it becomes a constant.
Although the mortality at each time interval is referred to as \textit{endpoint} in the ecotoxicology literature, we use \textit{outcome} of the experiment to avoid confusion.
Table \ref{tab:endpoints} shows the typical outcomes and their proportion within the effects data. 
To give a good indication of the toxicity to a species, these experiments 
need
to be repeated with increasing concentrations until the mortality reaches $100\%$. However, this is time consuming and is generally not done (\textit{sola dosis facit venenum}). Hence, some compounds may appear more toxic than others due to limited experiments. Thus, when evaluating prediction models, (higher values of) recall are preferred over precision.

Risk assessment methods require large amounts of effect data to efficiently
predict
long term risk for the ecosystems. The data must cover a minimum of the chemicals found when analysing water samples from the ecosystem, along with covering species present in the ecosystem. This leads to a immense search space that is close to impossible to encompass in its entirety.
Thus, it is essential to extrapolate from known to unknown combinations of chemical-species and suggest to the lab (ranked) effect hypothesis.
The state-of-the-art within effect prediction are quantitative structure–activity relationship models (QSARs). These models have shown promising results for use in risk assessment, \eg \cite{Pradeep2016}. 
However, QSARs have limitations with regard the coverage of compounds and species. 
These models use some chemical properties, but they usually only consider  one or few species at a time.
%
In this work we contribute with an alternative approach based on knowledge graph embeddings where the knowledge graph provides a global and integrated view of the domain.

Currently, the NIVA RAdb is under redevelopment, giving opportunities to include sophisticated effect prediction approaches, like the one presented in this paper, as a novel module for improving domain wide regulatory risk assessment.

\begin{figure}[t]
     \centering
     \begin{tikzcd}
                                            & ECOTOX \arrow[rd] \arrow[d] \arrow[ld, "Split" description] &                                                      \\
Species \arrow[dd, "Transform" description] & Effects \arrow[d]                                           & Compounds \arrow[dd]                                 \\
                                            & Map \arrow[d]                                               &                                                      \\
Alignment (LogMap) \arrow[ru] \arrow[r]     & \colorbox{red}{\textbf{TERA-KG}}                                                    & Alignment (Wikidata) \arrow[lu]                      \\
                                            &                                                             &                                                      \\
NCBI \arrow[uu, "Transform" description]    & ChEBI \arrow[uu, "SPARQL" description]                     & PubChem \arrow[luu, "Import" description] \arrow[uu]
\end{tikzcd}
     \caption{Data sources in the TERA knowledge graph. Compound classification is available from PubChem. Chemical class hierarchy comes from the ChEMBL SPARQL endpoint. Compound literals are gathered from PubChem REST API and transformed into triples. ECOTOX and PubChem identifiers are aligned using the Wikidata SPARQL endpoint. ECOTOX and NCBI  taxonomies are aligned using LogMap.}
     \label{fig:datasources}
\end{figure}
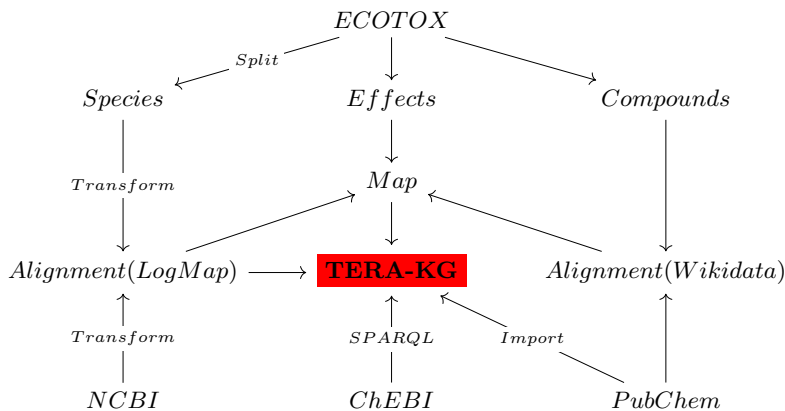

\section{A knowledge graph for toxicological effect data}

Risk assessment involves different data sources and laboratory experiments as shown in Figure \ref{fig:niva-pipeline}.
In this section we describe the relevant datasets and their integration to create the
\textit{Toxicological Effects and Risk Assessment} (TERA) knowledge graph (see Figure \ref{fig:datasources}).

\subsection{The ECOTOX database}
We rely on the ECOTOXicology database (ECOTOX) \cite{ecotox}. 
ECOTOX consists of $\sim 930k$ tests (or experiments) derived from the literature. Currently, an ECOTOX test considers the effect of one of $\sim 12k$ chemicals on one of $\sim 13k$ species. Which implies that less than $1\%$ of compound-species pairs have been tested. The effect is categorised in one of a plethora of predefined outcomes. For example, the $LC50$ outcome implies lethal concentration for $50\%$ of the test population. Table \ref{tab:endpoints} shows the most frequent outcomes in ECOTOX.

\begin{table}[t]
    \centering
    \begin{tabular}{c|c|c|c}
        test\_id & reference\_number & test\_cas & species\_number \\ \hline 
        $1068553$ & $5390$ & $877430$ (2,6-Dimethylquinoline) & $5156$ (Danio rerio) \\
        $2037887$ & $848$ & $79061$ (2-Propenamide) & $14$ (Rasbora heteromorpha)
    \end{tabular}
    \begin{tabular}{c|c|c|c|c}
        result\_id & test\_id & endpoint & conc1\_mean & conc1\_unit \\ \hline 
        $98004$ & $1068553$ & $LC50$ & $400$ & $mg/kg$ diet \\
        $2063723$ & $2037887$ & $LC10$ & $220$ & $mg/L$
    \end{tabular}
    \medskip
    \caption{ECOTOX database entry examples.}
    \label{tab:ecotox_ex}
\end{table}

\begin{figure}[t]
    \centering
    \includegraphics[width=0.75\textwidth]{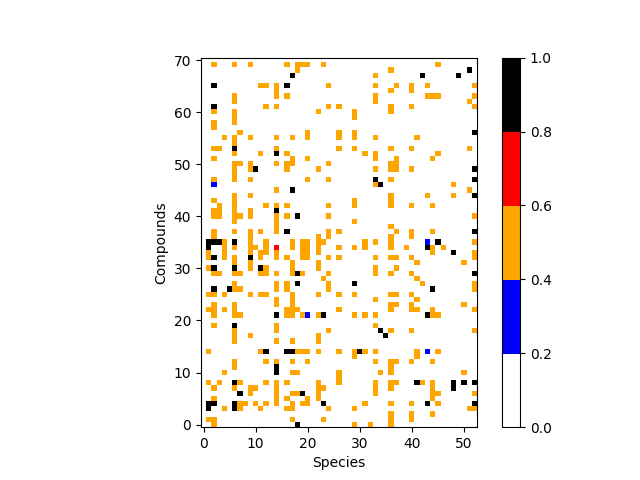}
    \caption{ECOTOX effects data. $x$ and $y$-axis represent individual species and chemicals sorted by similarity. Similarities are given by Equations \eqref{eq:adjacency} and \eqref{eq:similarity} in Section \ref{sect:baseline}. \ie chemicals $c_i\in C$ are indexed such that $S_{0,1}>S_{1,2}>\dots >S_{n-1,n}$. 
    Showing only chemicals and species that are involved in 25 or more experiments. 
    Values relate to mortality rate of the test population, \ie LC50 corresponds~to~$0.5$.}
    \label{fig:effects}
\end{figure}

Table \ref{tab:ecotox_ex} contains an excerpt of the ECOTOX database. 
ECOTOX includes information about the compounds and species used in the tests. 
This information, however, is limited and additional (external) resources are required to complement ECOTOX.

The number of outcomes per compound and species varies substantially. For example, there are 1,881 experiments where the compound used is \textit{sulfuric acid}, and 9,436 experiments where \textit{Pimephales promelas} (fathead minnow) is the test species. 
The median number of experiments per chemical and species are $3$ and $6$, respectively. Figure \ref{fig:effects} visualises a subset of the outcomes, here the zero values are either no effect or missing. This figure shows certain features of the data, \eg that compounds are more diversely used than species and that compound similarity is closely correlated to effects with regards to a species.

Currently, the ECOTOX database in used in risk assessment as reference data when calculating risk for a ecosystem. Essentially, comparing the reference and the observed chemical concentrations (per species). Since most compounds have multiple experiments per species, the mean and standard deviation of risk to a species can be calculated. However, if there is only one experiment for a compound-species pair we cannot calculate a standard deviation, such that the risk assessment is featureless. Therefore, estimating new effects is important to represent the natural variability of the effect data.

\subsection{Dataset integration into the TERA knowledge graph}

Figure \ref{fig:datasources} shows the different datasets and their transformation that contribute in the creation of the TERA knowledge graph. For example Triples \textit{(vii)}-\textit{(ix)} in Table~\ref{tab:triples} have been created from the ECOTOX effect data.

Each compound in the ECOTOX  effect data has a identifier called CAS Registry Number assigned by the Chemical Abstracts Service. The CAS numbers are proprietary, however, Wikidata \cite{wikidata2014} (indirectly) encodes mappings between CAS numbers and open identifiers like \textit{InChIKey}, a 27 character hash of the International Chemical Identifier (InChI) that encodes the chemical information in a unique manner. Hence, other datasets, such as PubChem \cite{pubchem}, can be used to gather chemical features and classification of compounds. PubChem is already available as a knowledge graph and can be imported directly. However, the PubChem hierarchy only contains permutations of compounds. 
To create a full taxonomy for the chemical data, we use the ChEMBL SPARQL endpoint to extract the classification (provided by the ChEBI ontology \cite{ChEBI}) for the relevant PubChem compounds. For example Triples \textit{(v)} and \textit{(vi)} in Table \ref{tab:triples} come from the integration with PubChem and ChEMBL.

\begin{table}[t]
    \centering
    \begin{tabular}{c c c c} 
        \texttt{\#} & \texttt{subject} & \texttt{predicate} & \texttt{object} \\ \hline
        \texttt{(i)} & \texttt{ecotox:group/Worms} & \texttt{owl:disjointWith} & \texttt{ecotox:group/Fish} \\
        \texttt{(ii)} & \texttt{ncbi:division/2} & \texttt{owl:disjointWith} & \texttt{ncbi:division/4} \\ 
        \texttt{(iii)} & \texttt{ecotox:taxon/34010} & \texttt{rdfs:subClassOf} & \texttt{ecotox:taxon/hirta} \\
        \texttt{(iv)} & \texttt{ncbi:taxon/687295} & \texttt{rdfs:subClassOf} & \texttt{ncbi:taxon/513583} \\\hline
        \texttt{(v)} & \texttt{compound:CID10198308} & \texttt{rdf:type} & 
        \texttt{obo:CHEBI\_134899} \\
         \texttt{(vi)} & \texttt{compound:CID10198308} & \texttt{pubchem:formula} & \texttt{``$C_7H_6O_6S$''} \\\hline
        \texttt{(vii)} & \texttt{ecotox:chemical/115866} & \texttt{ecotox:affects} & \texttt{ecotox:effect/001} \\
        \texttt{(viii)} & \texttt{ecotox:effect/001} & \texttt{ecotox:species} & \texttt{ecotox:taxon/26812} \\
        \texttt{(ix)} & \texttt{ecotox:effect/001} & \texttt{ecotox:endpoint} & \texttt{LC50}\\\hline
        \texttt{(x)} & \texttt{ecotox:taxon/33155} & \texttt{owl:sameAs} & \texttt{ncbi:taxon/311871}
    \end{tabular}
    \medskip
    \caption{Example triples from the TERA knowledge graph}
    \label{tab:triples}
\end{table}

\medskip
\noindent
\textbf{Aligning ECOTOX and NCBI}. 
The species lineage in ECOTOX is not complete and therefore this (missing) information has been complemented with the NCBI taxonomy \cite{ncbi}, a curated classification of all of the organisms in the public sequence databases (around $10\%$ of the species on Earth). 
The tabular data provided for the ECOTOX species and the NCBI taxonomies has been transformed into subsumptions and disjointness triples (see first four triples in Table \ref{tab:triples}). Leaf nodes are treated as instance entities. 

Since there does not exist a complete and public alignment between ECOTOX species and the NCBI Taxonomy, we have used the LogMap \cite{logmap2011,logma_ecai2012} ontology alignment systems to index and align the ECOTOX and NCBI vocabularies.
ECOTOX currently only provides a subset of the mappings via its web search interface. We have gathered a total of $929$ ground truth mappings for validation purposes.
The lexical indexation provided by LogMap left us with 5,472 possible NCBI entities to map to ECOTOX (we focus only on instances, \ie leaf nodes). LogMap identified 4,681 (instance) mappings to ECOTOX ($\sim 40\%$ of its entities) covering all $929$ mappings from the (incomplete) ground truth, thus, an estimated recall of $100\%$. The mappings computed by LogMap have been included to the TERA knowledge graph as additional equivalence triples (see Triple \textit{(x)} in Table \ref{tab:triples} as example). 

\section{Effect prediction models}

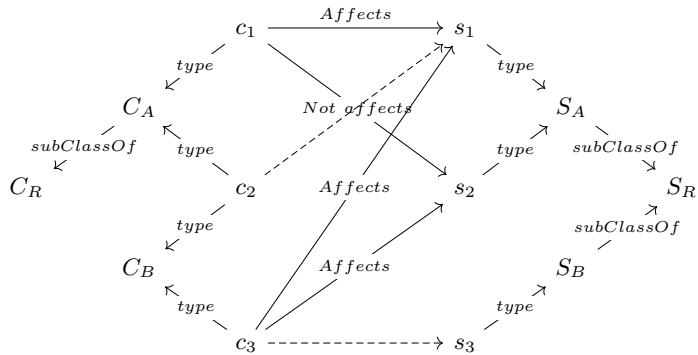
\begin{figure}[tb]
     \centering
     \begin{tikzcd}
    &                                          & c_1 \arrow[ld, "type" description] \arrow[rrr, "Affects"] \arrow[rrrdd, "Not\ affects" description]                                &  &  & s_1 \arrow[rd, "type" description] &                                          &     \\
    & C_A \arrow[ld, "subClassOf" description] &                                                                                                                                    &  &  &                                    & S_A \arrow[rd, "subClassOf" description] &     \\
C_R &                                          & c_2 \arrow[ld, "type" description] \arrow[lu, "type" description] \arrow[rrruu, dashed]                                            &  &  & s_2 \arrow[ru, "type" description] &                                          & S_R \\
    & C_B                                      &                                                                                                                                    &  &  &                                    & S_B \arrow[ru, "subClassOf" description] &     \\
    &                                          & c_3 \arrow[lu, "type" description] \arrow[rrruu, "Affects" description] \arrow[rrruuuu, "Affects" description] \arrow[rrr, dashed] &  &  & s_3 \arrow[ru, "type" description] &                                          &    
\end{tikzcd}
     \caption{The effect prediction problem. 
     Lowercase $s_j$ and $c_i$ are instances of species and compounds, while uppercase denote classes in the hierarchy. Solid lines are observations and dashed lines are to be predicted. 
     \ie does $c_2$ affect $s_1$?}
     \label{fig:affects}
\end{figure}

In this section we introduce the selected machine learning models to solve the effect prediction problem shown in Figure \ref{fig:affects}. We use the known effects, denoted as \textit{Affects} and \textit{Not affects} in the figure, to predict whether or not new proposed chemical-species pairs are \textit{true} (Affects) or \textit{false} (Not affects).\footnote{The models are implemented with Keras~\cite{chollet2015keras}. Data and codes available from: \url{https://github.com/Erik-BM/NIVAUC}}

\medskip
\noindent
\textbf{Effect data sampling}. A balance between positive and negative effect data samples is desired, therefore, we choose outcomes in categories (refer to Table~\ref{tab:endpoints}): NOEL, LCp, LDp, NR-LETH, and NR-ZERO (p ranges from $0$ to $100$). 
We are only concerned about the 
mortality rate in experiments, consequently, we treat LC* and LD* identically. In addition, NR-LETH is treated as LC100. For simplicity, we treat the effects as binary entities.
Hence, the outcome for a compound-species pair $c,s$ is defined as
\begin{align}
    f(c,s) = \begin{cases}
        1 & \text{if } (c,s) \in \text{LCp $\cup$ LDp $\cup$ NR-LETH} \\
        0 & \text{if } (c,s) \in \text{NOEL $\cup$ NR-ZERO.}
    \end{cases}
\end{align}

For example, according to Figure \ref{fig:affects}, $f(c_1,s_1)=1$ (\ie $c_1$ affects $s_1$) and $f(c_1,s_2)=0$ (\ie $c_1$ does not affects $s_1$), while $f(c_2,s_1)$ is unknown and thus a prediction is required for this chemical-species pair.

\medskip
\noindent
\textbf{Knowledge graphs.}
We rely on the TERA knowledge graph (see excerpts in Table \ref{tab:triples} and Figure \ref{fig:affects}) to feed the knowledge graph embedding algorithms. For simplicity we discard the ECOTOX species entities that have not a correspondence to NCBI. Note that we currently do not consider literals.


\subsection{Baseline model ($M_1$)}
\label{sect:baseline}
This (baseline) prediction model is based on the current prediction method used at NIVA. The basic idea of this method is to find the nearest-neighbour from the observed samples. In this context, the nearest neighbours are defined by hierarchy distance for species and similarity for compounds. Therefor, we first define a adjacency matrix for the taxonomy and a similarity matrix for compounds.
\begin{equation}
   A_{i,j} = \frac{1}{|P(s_i,r)|+|P(s_j,r)|-2|P(s_i,r)\cap P(s_j,r)|+1} 
   \label{eq:adjacency}
\end{equation}
where $r$ is the taxonomy root, $P(x,r)$ is the classes in the path from $x$ to $r$, and $\left|\cdot\right|$ denotes the cardinality.
One basic approach to calculate the chemical similarity is using the Jaccard index of the binary fingerprints of the compounds~\cite{doi:10.1002/qsar.200330831}. 
Hence, the similarity matrix is defined as 
\begin{equation}
    S_{i,j} = J(c_i,c_j) = \frac{|(F_i)_2 \cap (F_j)_2|}{|(F_i)_2 \cup (F_j)_2|}
    \label{eq:similarity}
\end{equation}

We define a matrix $E\in \mathbb{R}^{|C|\times |T|}$, where $C$ and $T$ denote the set of compounds and species respectively.
$E$ contains all the observed effects (training set):
\begin{align}
    E_{i,j} &= \begin{cases}
        1 & \text{if } (c_i,\text{ affects, } s_j)  \\
        0 & \text{else}
    \end{cases}
\end{align}

We can then make the prediction with $A$, $S$, and $E$
, as shown in Algorithm~\ref{alg:exhaustive}. 
The algorithm terminates when $t_{max}$ neighbours are visited or $p>0$.

\begin{algorithm}[t]
\SetAlgoLined
\KwInput{$E$, $A$, $S$, $c_i$, $s_j$}
\KwOutput{$p$, effect prediction for $c_i, s_j$}
 $i^\prime,j^\prime \gets i,j$\;
 $t_1 \gets t_{max}$\;
 $p\gets E_{i,j}$ \tcp*{0 if no overlap between train and test}\
 \While{$t_1>0$}{
    $i^\prime \gets \argmax_{k\neq i} S_{i,k}$\tcp*{find index of most similar compound}\
    $A^\prime \gets A$; $t_2 \gets t_{max}$\tcp*{copy A and reset counter}\
    reset $j$\tcp*{reset to $j$ in input $s_j$}
    \While{$t_2 > 0$}{
    $j^\prime \gets \argmax_{k\neq j} A^\prime_{j,k}$\tcp*{index of closest specie}\
    $p\gets \max{(p,E_{i^\prime,j^\prime})}$\tcp*{update prediction}\
    $A^\prime_{j,j^\prime} \gets 0$\tcp*{set seen indices to zero}\ 
    $t_2 \gets t_2 -1$\;
    \lIf{$p>0$}{\Return $p$}
    $i,j\gets i^\prime,j^\prime$\tcp*{update}\
    }
    $S^\prime_{i,i^\prime} \gets 0$\tcp*{set seen indices to zero}\
    $t_1 \gets t_1 - 1$\;
 }
 \Return $p$\;
 \caption{
Baseline prediction model algorithm ($M_1$).
 }\label{alg:exhaustive}
\end{algorithm}


\subsection{Multilayer perceptron ($M_2$)}
Our second prediction model is a Multilayer perceptron (MLP) network with $n$ hidden layers. The model can be~expressed~as:
\begin{align}
    \vec{y}^0 &= \left[\vec{e}_c, \vec{e}_s\right] \\
    \vec{y}^t &= ReLu(\vec{y}^{t-1} W_t + \vec{b}_t) \\
    \hat{y} &= \sigma(\vec{y}^n W_n + \vec{b}_n)
\end{align}
where $t = 1,2,...,n$. $\left[\cdot, \cdot \right]$ denotes vector concatenation. $ReLu$ is the rectifier function and $\sigma$ is the logistic sigmoid function. $W_t$ are the weight matrices and $b_t$ are the biasses for each layer. $\vec{e}_c,\vec{e}_s \in \mathbb{R}^k$ are the embedded vectors of $c$ and $s$. For example $\vec{e}_c$ is defined as
\begin{align}
    \vec{e}_c = \vec{\delta}_c W_C
\end{align}
where $\vec{\delta}_c$ is the one-hot encoded vector for entity $c$,
$W_C \in \mathbb{R}^{|C|\times k}$ is an embedding transformation matrix to learn. 

A dropout layer is stacked after each hidden layer to prevent the network from overfitting.
The model is optimised using \texttt{ADAGRAD} \cite{Duchi:2011:ASM:1953048.2021068} with the following log loss function:
\begin{align}
    L(\vec{y},\vec{\hat{y}}) = - \frac{1}{N}\sum_{i=1}^N\left[ y_i \log{\hat{y}_i} + (1-y_i)\log{(1-\hat{y}_i)}\right]
\end{align}

\subsection{Knowledge graph (KG) embedding and MLP ($M_2^\star$)}
We have extended the MLP model ($M_2$) by feeding it with the TERA KG-based embeddings of $c$ (\ie the chemical) and $s$ (\ie the species), which encode the information of the taxonomy and compound hierarchies, among other semantic relationships.
%
Note that the TERA knowledge graph also includes similarity triples about compounds. These triples represent pairs of compounds $c_i$ and $c_j$ where their similarity $S_{i,j}$ (as in Equation \ref{eq:similarity}) is above a threshold $\phi$.


The embeddings are learned by applying the scoring function from one of DistMult \cite{Yang2015EmbeddingEA}, HolE \cite{DBLP:journals/corr/NickelRP15}, and TransE \cite{NIPS2013_5071}. TransE was selected as it provides a very intuitive model. DistMult was included as it has shown state-of-the-art performance (\eg \cite{DBLP:journals/corr/KadlecBK17}), while HolE was considered as it also encodes directional relations.
The score function for DistMult is defined as 
\begin{align}
    S_D(s,p,o) = \sigma( \vec{e}_s^T W_p \vec{e}_o ), \ W_p = diag(\vec{e}_p)
\end{align}
HolE uses a circular correlation score function, defined by
\begin{align}
    S_H(s,p,o) = \sigma(\vec{e}_r^T [\vec{e}_s \star \vec{e}_o]), \ \vec{e}_s \star \vec{e}_o = \mathcal{F}^{-1}[\overline{\mathcal{F}(\vec{e}_s)}\odot \mathcal{F}(\vec{e}_o)]
\end{align}
where $\mathcal{F}$ and $\mathcal{F}^{-1}$ are the Fourier transform and its inverse, $\overline{x}$ is the elementwise complex conjugate, $\odot$ denotes the Hadamard product. 
The final method is TransE, which has the score function
\begin{align}
    S_T(s,p,o) = ||\vec{e}_s + \vec{e}_p - \vec{e}_o||
\end{align}
where $||\vec{x}||$ is the norm of $\vec{x}$. $\vec{e}_s$, $\vec{e}_p$ and $\vec{e}_o$ are the vector representation for the subject, predicate and object of a triple, respectively.

DistMult and HolE optimises for a score of $1$ for positive samples and $0$ for negative samples. Moreover, TransE scores positive samples as $0$ and with no upper bound for negative samples. We modify the TransE score function to $S_T^{\prime} = \tanh{(1/S_T)}$, such that $\lim_{S_T \to 0} S_T^{\prime} = 1$ and $\lim_{S_T \to \infty} S_T^{\prime} = 0$, to avoid modifying the labels.

The embeddings are used in the same network as the $M_2$ model. We train the embeddings and the classifier simultaneously using log loss and \texttt{ADAGRAD}. Training simultaneously will optimise the embeddings with regards to both the knowledge graph triples and the classifier loss.

\section{Effect prediction evaluation}

\noindent
\textbf{Sampling.}
We split the effect data $50\%$/$50\%$ for train/test. 
To prevent test set leakage, 
those training inputs that appear in the test set are removed,
resulting in a $70\%/30\%$ split.
$M_2^\star$ can be trained with the entirety of the knowledge graph, which is ignored under effect prediction. The negative knowledge graph samples are generated by randomly re-sampling subject and object of a true sample, while maintaining the distribution of predicates. We generate four negative samples per positive sample.

\medskip
\noindent
\textbf{$\mathbf{M_1}$ model settings.}
We tested the performance of $M_1$ with 6 choices of nearest neighbour ($5$, $10$, $20$, $30$, $40$, $50$). In addition to Algorithm \ref{alg:exhaustive}, we tested 
an alternative
technique for iterating over the data. However, 
Algorithm \ref{alg:exhaustive} yielded better results.
The most balanced results were found when using $30$ neighbours. When using more than $30$ neighbours recall increases, but accuracy and precision suffer from
a considerable decrease since the use of more neighbours increases the false positive rate.


\begin{table}[t]
    \centering
    \begin{tabular}{r|c|c|c|c|c}
         & $M_1$ ($t_{max} = 30$) & $M_2$ & $M_2^\star$ ($S_T^\prime$) & $M_2^\star$ ($S_D$) & $M_2^\star$ ($S_H$)  \\ \hline
         Accuracy & $0.58$ & ${0.82}$ & $\mathbf{0.83}$ & $\mathbf{0.83}$ & $\mathbf{0.83}$\\
         Precision & $0.47$ & $\mathbf{0.76}$ & $ {0.75}$ & $ \mathbf{0.76}$ & ${0.73}$\\
         Recall & ${0.80}$ & $0.78$ & ${0.84}$ & $ {0.82}$ & $\mathbf{0.87}$\\
         $F_1$ score & $0.59$ & $0.77$ & $\mathbf{0.79}$ & $\mathbf{0.79}$ & $\mathbf{0.79}$\\
         $F_{\beta=2}$ score & $0.70$ & $0.79$ & ${0.82}$ & ${0.81}$ & $\mathbf{0.84}$\\
         AUC & $-$ & ${0.90}$ & $\mathbf{0.91}$ & $ \mathbf{0.91}$ & $\mathbf{0.91}$\\ \hline
         Accuracy & $0.56\pm 0.01$ & $\mathbf{0.81 \pm 0.02}$ & $\mathbf{0.81\pm 0.02}$ & $\mathbf{0.81 \pm 0.01}$ & $ \mathbf{0.81\pm 0.02}$\\
         Precision & $0.55\pm 0.01$ & ${0.79 \pm 0.04}$ & $\mathbf{0.80\pm 0.04}$ & $ {0.78 \pm 0.03}$ & $ {0.79\pm 0.03}$\\
         Recall & ${0.76\pm 0.03}$ & $0.84 \pm 0.08$ & ${0.83\pm 0.08}$ & $\mathbf{0.87 \pm 0.05}$ & $ {0.86\pm 0.02}$\\ 
         $F_1$ score & $0.65\pm 0.01$ & ${0.81\pm 0.03}$ & ${0.81\pm 0.03}$ & $\mathbf{0.82\pm 0.01}$ & ${0.82\pm 0.01}$\\
         $F_{\beta=2}$ score & $0.72\pm 0.02$ & $0.83\pm 0.06$ & ${0.82\pm 0.06}$ & $\mathbf{0.85\pm 0.03}$ & ${0.84\pm 0.01}$\\
         AUC & $-$ & $\mathbf{0.89 \pm 0.01}$ & ${0.88\pm 0.01}$ & $ \mathbf{0.89 \pm 0.01}$ & $ \mathbf{0.89\pm 0.02}$
    \end{tabular}
    \medskip
    \caption{Performance of the prediction models. $M_2^\star$ ($S_T^\prime$),  $M_2^\star$ ($S_D$) and $M_2^\star$ ($S_H$) stand for the MLP prediction models using  TransE, DistMult, and HolE embedding models, respectively. \textit{Above line}: ensemble averages of $10$ clean tests. \textit{Below line}: 10 fold cross validation on training set with standard deviation.}
    \label{tab:metrics}
\end{table}

\medskip
\noindent
\textbf{$\mathbf{M_2/M_2^\star}$ model settings.}
The embedding dimension used in $M_2$ and $M_2^\star$ was based on a search among sizes $16$, $64$, $128$ and $256$. We found no difference between these parameters for $M_2$, therefor, $16$ is chosen to aid faster training. $M_2^\star$ used a larger amount of entities and needs a larger embedding space to capture the features of the data. The performance plateaued at $128$, hence, this was chosen.
The models ($M_2$,$M_2^\star$) were trained until the loss stops improving for 5 iterations. For $M_2^\star$ we used different loss weights for the embeddings and the effect predictor. These weights were chosen such that the embeddings and effects are learned at similar rates. 
DistMult and HolE used $0.5$ and $1.0$ as loss weights for embeddings and effects models, respectively, while TransE used equal weights. We used a dropout rate of $0.2$ and a similarity threshold of $0.5$. 
Note that in $M_2^\star$ we simultaneously train the embedding models and the effect predictor.
We perform 
\begin{inparaenum}[\it (i)]
\item 10 fold cross validation on the training set, and \item a clean test on the unseen test set. This test consist of a ensemble of 10 models trained on the training set, each with a new set of random negative knowledge graph samples. We used an ensemble to limit the impact the random negative samples has on the results.
\end{inparaenum}

\begin{center}
\begin{figure}[t!]
    \centering
    \begin{subfigure}[b]{0.87\textwidth}
        \includegraphics[width=\textwidth]{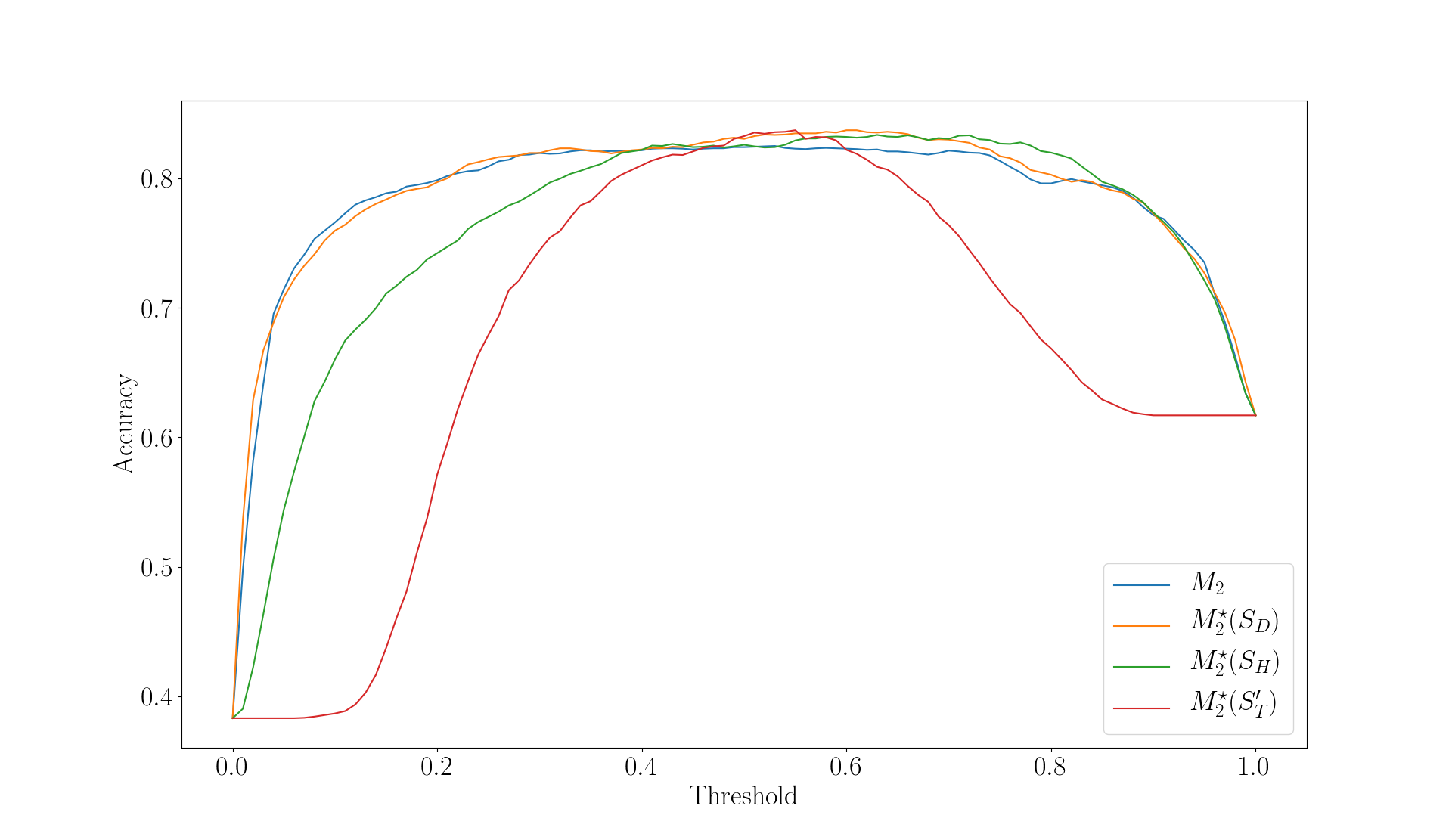}
        \caption{Accuracy for the $M_2$ and $M_2^\star$ prediction models.}
        \label{fig:accuracy}
    \end{subfigure}
    \centering
    \begin{subfigure}[b]{0.87\textwidth}
        \includegraphics[width=\textwidth]{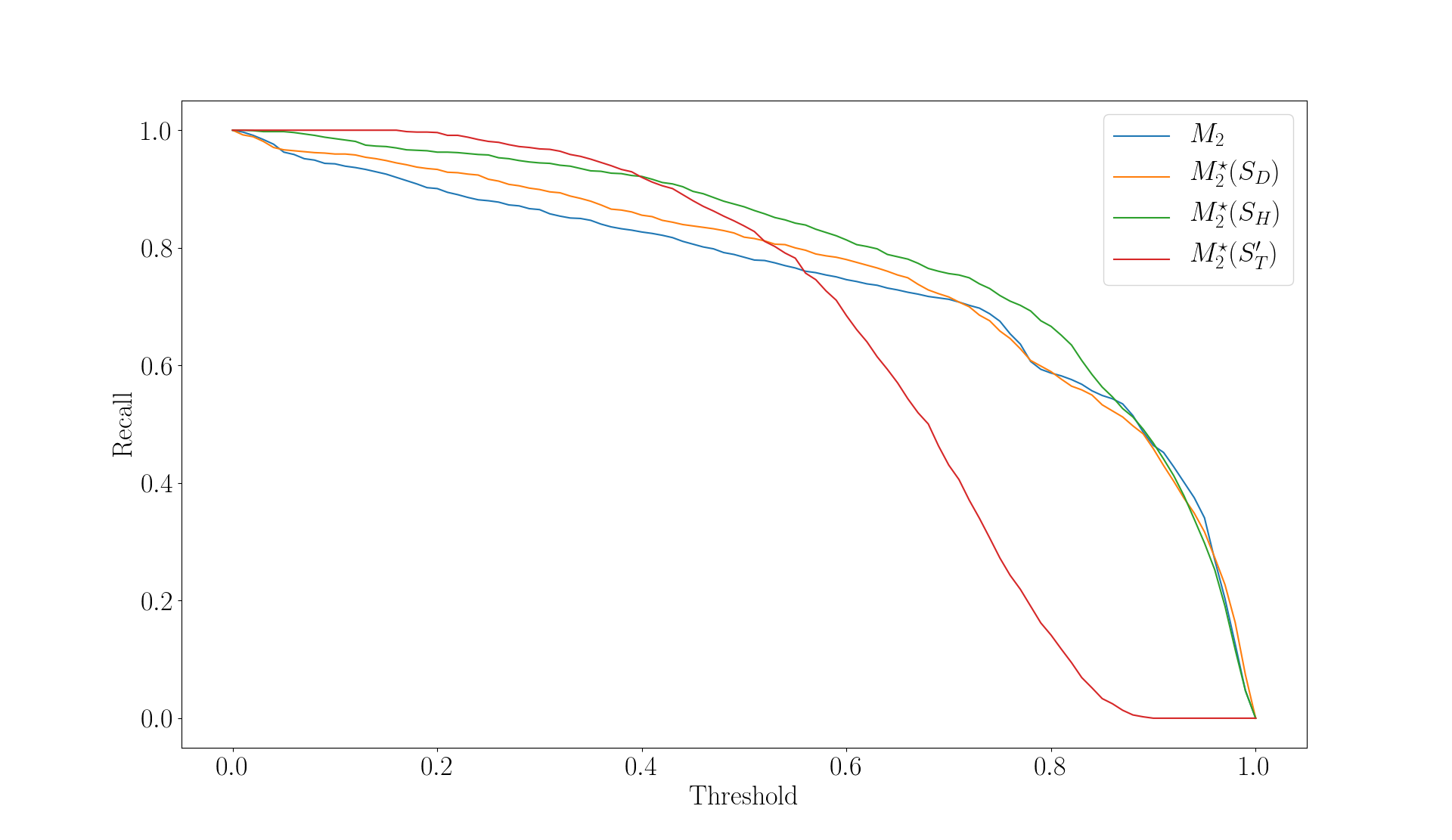}
        \caption{Recall for the $M_2$ and $M_2^\star$ prediction models.}
        \label{fig:recall}
    \end{subfigure}
    \caption{Accuracy and Recall for the $M_2$ and $M_2^\star$ models with various thresholds.}\label{fig:acc-recall}
\end{figure}
\end{center}

\medskip
\noindent
\textbf{Evaluation}. Figures \ref{fig:accuracy} and \ref{fig:recall}  and Table \ref{tab:metrics} show the results of the conducted evaluation for the five effect prediction models. Figures \ref{fig:accuracy} and \ref{fig:recall} visualise the impact on accuracy and recall with different thresholds on the $M_2$-$M_2^\star$ prediction scores, while Table \ref{tab:metrics} presents the relevant evaluation metrics with a threshold~of~$0.5$ for $M_2$-$M_2^\star$ and $30$ neighbours for $M_1$.
The results can be summarised as follows:
\vspace{-0.2cm}
\begin{enumerate}[\it (i)]

\item 
$M_1$ is only slightly better than random choice, as the prior binary output distribution is $0.59$ and $0.41$. 
Thus it would not be appropriate for predicting effects. The false positive rate is also high, hence, $M_1$ would not be practical to use as a recommendation system.

\item $M_2$ is considerably better than $M_1$ 
and has balance between precision and recall. We suspect that this balance is due to random choice when the model has not previously seen a chemical or species.
\ie a prediction close to the decision boundary when an input is unseen will maintain the false negative/positive proportion, hence good for accuracy, not necessary for giving (interesting) recommendations to the laboratory. 

\item
Introducing the background knowledge to $M_2$, in the form of KG embeddings gives higher recall, without loosing accuracy. In contrast to $M_2$, $M_2^\star$ is more uncertain when unseen combinations are presented to the model (\textit{in dubio pro reo}).
Therefore, $M_2^\star$ is better suited to giving recommendations for cases where there is limited information about the chemical and the species in the effect data.

\item 
The best results in terms of recall, when using a threshold of $0.5$ (see Table~\ref{tab:metrics}), are obtained
by $M_2^\star$ with the embeddings provided by HolE 
($9$~points higher than the $M_2$).

\item 
As shown in Figures \ref{fig:accuracy} and \ref{fig:recall}, lowering the decision threshold ($0.30$) would yield a higher recall ($0.90$) for the DistMult-based model, while maintaining the accuracy. 
TransE and HolE-based models have higher recall ($0.97$ and $0.94$) at decision threshold $0.30$, however, this comes at a cost of reduction in accuracy ($0.74$ and $0.79$).

\item 
The highest overall $F_{\beta=2}$ score is $0.87$, and is shared by all $M_2^\star$ models, albeit, at different decision boundaries, $0.34$, $0.14$ and $0.31$ for models with TransE, DistMult, and HolE embeddings, respectively.
\end{enumerate}

\section{Discussion and future work}

We have created a knowledge graph called TERA that aims at covering the knowledge and data relevant to the ecotoxicological domain. We have also implemented a proof-of-concept prototype for ecotoxicological effect prediction based on knowledge graph embeddings. The obtained results are encouraging,
showing the positive impact of using knowledge graph embedding models and the benefits of having an integrated view of the different knowledge and data sources.

\medskip
\noindent
\textbf{Knowledge graph.} The TERA knowledge graph is by itself an important contribution to NIVA. TERA integrates different knowledge and data sources and aims at providing an unified view of the information relevant to the ecotoxicology and risk assessment domain. At the same time the adoption of a RDF-based knowledge graph enables the use of 
\begin{inparaenum}[\it (i)]
\item an extensive range of Semantic Web infrastructure that is currently available (\eg reasoning engines, ontology alignment systems, SPARQL query engines), and
\item state of the art knowledge graph embedding strategies.
\end{inparaenum}

\medskip
\noindent
\textbf{Prediction models.} The obtained predictions are promising and show the validity of the selected models in our setting and the benefits of using the TERA knowledge graph. As mentioned before, we favour recall with respect to precision. One the one hand, false positives are not necessarily harmful, while overlooking the hazard of a chemical may have important consequences. On the other hand, due to the limited experiments in terms of concentration (\ie effect data may not be complete), some chemicals may look less toxic than others while they may still be hazardous.


\medskip
\noindent
\textbf{Value for NIVA.} 
The conducted work falls into one of the main research lines of NIVA's Computational Toxicology Program (NCTP) to enhance the generation of hypothesis to be tested in the laboratory \cite{setac2019}. Furthermore, the data integration efforts and the construction of the TERA knowledge graph also goes in line with the vision of NIVA's section for Environmental Data Science. The availability and accessibility of the best knowledge and data will enable optimal decision making. 


\medskip
\noindent
\textbf{Novelty.} Knowledge graph embedding models have been applied in general purpose link discovery and knowledge graph completion tasks \cite{KGE_survey_2017}. They have also attracted the attention in the biomedical domain to find, for example, candidate genes for a disease, protein-protein interactions or drug-target interactions (\eg \cite{DBLP:journals/bioinformatics/AlshahraniKMKQH17,DBLP:conf/semweb/AgibetovS18}).
However, we are not aware of the application of knowledge graph embedding models in the context of toxicological effect prediction.

\medskip
\noindent
\textbf{Future work.}
The main goal in the mid-term future is to integrate the TERA knowledge graph and the machine learning based prediction models within NIVA's risk assessment pipeline. In the near future, we intend to improve the current ecotoxicological effect prediction prototype and 
evaluate the suitability of more sophisticated models like Graph Convolutional Networks.
The TERA knowledge graph will also be extended with additional information about species (\eg interactions) and compounds (\eg target proteins) which is expected to enhance the computed embeddings and the effect predictions. 


\medskip
\noindent
\textbf{Resources.} The datasets, evaluation results, documentation and source codes are available from the following GitHub repository: \url{https://github.com/Erik-BM/NIVAUC}

\section*{Acknowledgements}

This work is supported by the grant 
272414 from the Research Council of Norway (RCN), the MixRisk project (RCN 268294), the AIDA project, 
The Alan Turing Institute under the EPSRC
grant EP/N510129/1, 
the SIRIUS Centre for Scalable Data Access (RCN 237889),
the Royal Society, EPSRC projects DBOnto, $\text{MaSI}^{\text{3}}$ and $\text{ED}^{\text{3}}$. 
We would also like to thank Martin Giese and Zofia C. Rudjord for their contribution in early stages of this project.


%
%
%
\bibliographystyle{splncs04}
 \bibliography{references}
\end{document}